\documentclass[journal]{IEEEtran}

\usepackage{graphicx}

\usepackage{tabularx}

\usepackage[table]{xcolor}
\usepackage{hhline}

\usepackage{blindtext}
\usepackage{enumitem}
\usepackage{graphicx}
\usepackage{algorithm}
\usepackage{algpseudocode}
\usepackage{subcaption}

\ifCLASSINFOpdf
\else
\fi
\begin{document}
%
\title{The 2017 AIBIRDS Competition}
%
%
%

\author{Matthew~Stephenson,
        Jochen~Renz,
	Xiaoyu~Ge,
	and~Peng~Zhang
\thanks{M. Stephenson, J. Renz, X. Ge and P. Zhang are with the Research School of Computer Science, Australian National University, Canberra, A.C.T. 0200, Australia, e-mail: (matthew.stephenson@anu.edu.au).}
}

\maketitle

\begin{abstract}

This paper presents an overview of the sixth AIBIRDS competition, held at the 26th International Joint Conference on Artificial Intelligence. This competition tasked participants with developing an intelligent agent which can play the physics-based puzzle game Angry Birds. This game uses a sophisticated physics engine that requires agents to reason and predict the outcome of actions with only limited environmental information. Agents entered into this competition were required to solve a wide assortment of previously unseen levels within a set time limit. The physical reasoning and planning required to solve these levels are very similar to those of many real-world problems. This year's competition featured some of the best agents developed so far and even included several new AI techniques such as deep reinforcement learning. Within this paper we describe the framework, rules, submitted agents and results for this competition. We also provide some background information on related work and other video game AI competitions, as well as discussing some potential ideas for future AIBIRDS competitions and agent improvements.



\end{abstract}


%
\IEEEpeerreviewmaketitle

\section{Introduction}

Over the past several years, many different AI competitions focused around video games have become extremely popular. Many of these competitions have yielded promising results and improvements for the wider AI community, and have been hosted at several major international conferences including CIG, AIIDE, IJCAI, ECAI, GECCO and FDG to name just a few. Whilst competitions and challenges centred around AI playing classic board games, such as chess with Deep Blue \cite{blue} and more recently Go with DeepMind's AlphaGo \cite{tog3}, have been incredibly popular and successful, video games typically provide a much more complex and challenging domain in which to interact. Developing agents (autonomous programs that can react intelligently to environmental inputs) that can successfully play popular and complex video games is a key area of research for AI. Video games provide a controllable and parameterised environment to work in, and the problems they pose are often very similar to those of the real-world \cite{agent1}. Most video games are designed to test the cognitive abilities of human players in one or multiple areas, which is precisely the problem we wish intelligent agents to solve. Physics-based puzzles are a great example of this as they not only require a fair amount of planning and knowledge reasoning, but also the type of physical reasoning required to play them is comparable to that needed for an agent to operate successfully in the real-world \cite{cite1}. Angry Birds is a popular video game that fits perfectly into this category.

Angry Birds is a physics-based simulation puzzle game, developed by Rovio Entertainment \cite{web}. It uses a sophisticated (at least in terms of other video game AI problems) physics engine to control the movement of certain objects and how they respond to the player's actions. Without getting too in-depth with its specific mechanics, players can only solve this game's levels by planning out a sequence of well-reasoned actions, taking into account the physical nature of the game's environment. This type of physical reasoning problem is very different to traditional games such as chess, as the exact attributes and parameters of various objects are often unknown. This means that it is very difficult to accurately predict the outcome of any action taken \cite{cite2} and the exact result of an action is only known for certain once it has been carried out. Even if the exact physics parameters of the world were available, the agent would have to simulate a potentially infinite number of possible actions due to the game's continuous state and action spaces. Developing intelligent agents that can play this game effectively has been an incredibly complex and challenging problem for traditional AI techniques to solve, even though the game is simple enough that any human player, or even a child, could learn it within a few minutes. Humans are naturally very good at predicting the result of a physical action based on visual information, while agents still struggle with this form of reasoning in unknown environments.

What makes this research on physics-based games such as Angry Birds so important, is that the exact same problems need to be solved by AI systems that are intended to interact successfully with the real-world. The ability to accurately estimate the consequences of a physical action based solely on visual inputs or other forms of perception is essential for the future of ubiquitous AI, and has huge real-world relevance and application. Any real-world AI system that cannot achieve this will likely result in many unintended outcomes which could potentially be dangerous to people. Angry Birds, as well as other physics-based games, provides a controlled and parametrised environment to experiment with new ideas and capabilities. 
It is particularly important for the development of such systems to integrate the areas of computer vision, machine learning, knowledge representation and reasoning, heuristic search, planning, and reasoning under uncertainty. Contributions or additions to each of these areas will help improve the overall performance of an agent, but combining effective solutions to all of these problems will be needed to develop a truly intelligent physical reasoning system. 

In this paper we present the description, entrants, results and conclusions for the sixth AIBIRDS competition. Participating competitors are tasked with developing an agent that can play and solve unknown Angry Birds levels. As previously mentioned, this competition was created as a means to promote the research and creation of intelligent agents that can reason and predict the outcome of actions in a physical simulation environment \cite{cite9}. 
During the competition, agents are required to play a set number of unknown levels within a given time, attempting to score as many points as possible in each level. The exact parameters of certain objects, as well as the current internal state of the game, are not directly accessible. Instead, information about the level is provided using a computer vision module which gives approximations of objects boundaries and location based on screenshots of the game screen, effectively meaning that an agent gets exactly the same input as a human player. Agents are required to solve these levels in real-time, and can attempt levels in any order and as many times as they like. Once the time limit has expired the maximum scores that an agent achieved for each solved level are summed up to give its final score. Agents are then ranked based on this value and after several rounds of elimination a winner is declared. The eventual goal of this competition is to design agents that can play new levels as well as, or better than, the best human players. Many of the previous agents that have participated in this competition employed a variety of techniques, including qualitative reasoning \cite{cite3}, internal simulation analysis \cite{cite5,cite4}, logic programming \cite{cite10}, heuristics \cite{cite11}, Bayesian inferences \cite{cite7,cite6}, and structural analysis \cite{cite8}.


Holding an AI competition has many advantages over traditional research methods, chief of which is that it informs members of the AI community who may not be aware of the problem about it. This is turn may encourage and motivate people to take part, perhaps inspiring them to try out their own methods and ideas to solve the problem. Competitions provide easy to use software and interfaces that can make a daunting and challenging task seem much more possible. Some participants may even be able to apply their existing algorithms to an entirely new problem they had not previously considered. Competitions also provide an effective way of comparing and benchmarking all of the currently existing algorithms. All submitted agents are evaluated using the same levels and rules, allowing for a fair and unbiased means of comparing them. This also provides opportunities for discussion and collaboration between researchers, and is a great way to get both industry specialists and other non-academics involved in this kind of work.


The remainder of this paper is organized as follows: Section II provides the background to this competition, including past AI video game competitions, a description of the Angry Birds game, and details on the related AIBIRDS level generation competition; Section III describes the competition itself, providing details on the naive agent that is provided to all entrants, as well as the rules and scoring procedure; Section IV contains descriptions of the ten agents submitted to this year's competition; Section V provides the results of the competition; Section VI discusses and interprets these results, providing some possible improvements for next year's agents and other potential benefits beyond the competition itself; Section VII presents our final conclusions and desired goals for future competitions.

\section{Background}

\subsection{Previous AI video game competitions}

Examples of popular AI competitions (both past and present) include the Mario AI Championship \cite{marioaicomp1,marioaicomp2,marioaicomp3}, the StarCraft AI Competition \cite{starcraftsurvey}, the Visual Doom AI Competition (ViZDoom) \cite{vizdoomcomp}, the Geometry Friends Game AI Competition \cite{tog10} and the Fighting Game AI Competition \cite{tog9}. The General Video Game AI (GVGAI) Competition has also run several tracks around developing agents for playing general video games. These include the single-player planning track \cite{gvgaicomp1}, the two-player planning track \cite{gvgai2player,togx3} and the learning track \cite{gvgaicomp2}. The AIBIRDS competition has itself been running since 2012 \cite{cite1,cite9}, with many advancements and improved agents being developed since its initial inception.  

\begin{figure}
    \centering
  \includegraphics[width=1.0\linewidth]{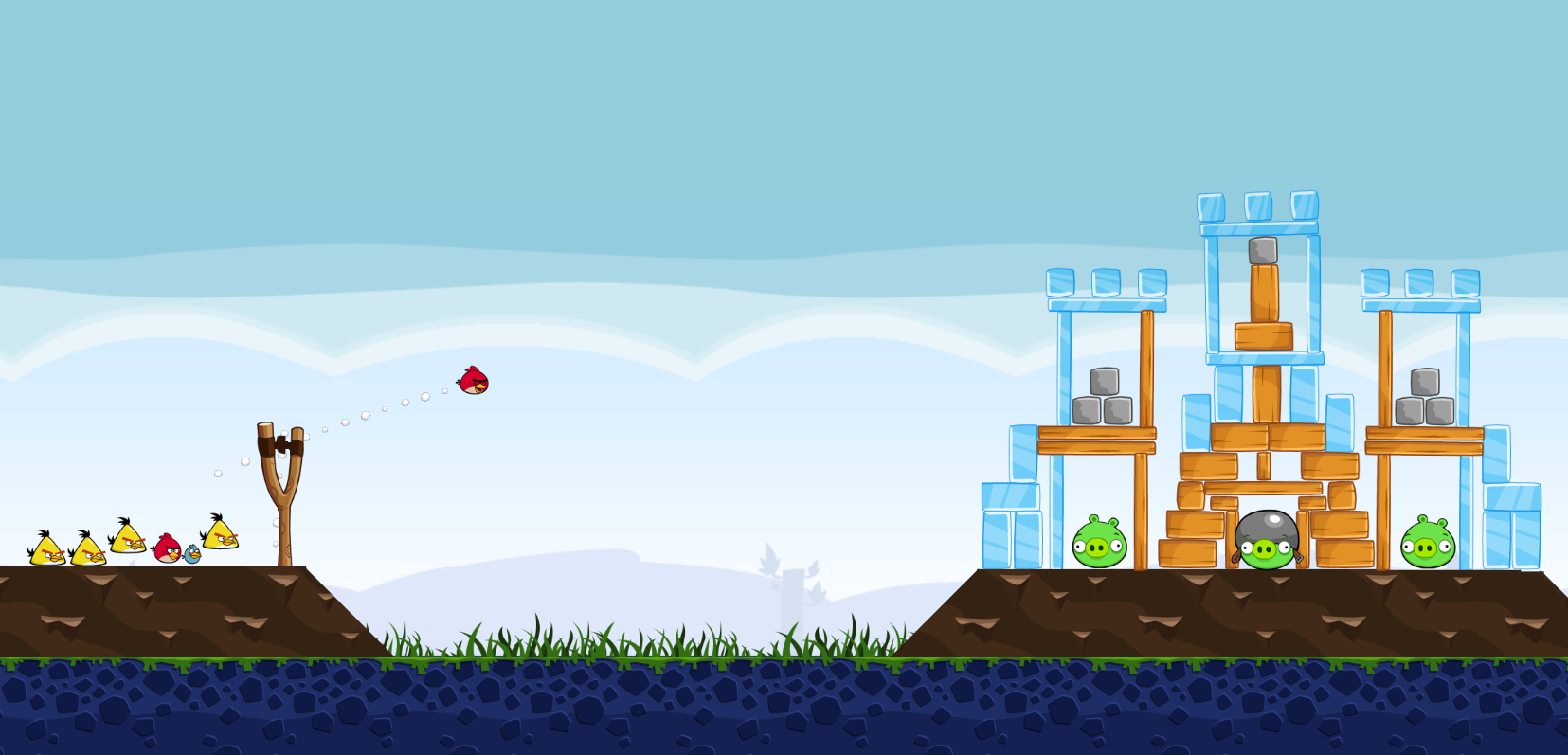}
  \caption{Screenshot of a level from the Angry Birds game.}
\end{figure}

\subsection{Angry Birds game}

Angry Birds is a popular physics-based puzzle game where in each level the player uses a slingshot to shoot birds at structures composed of blocks, with pigs placed within or around them \cite{web}. The player's objective is to kill all the pigs within a level using the birds provided. A typical Angry Birds level, as shown in Figure 1, contains a slingshot, birds, pigs and a collection of blocks arranged in one or more structures. All objects within the level have properties such as location, size, mass, friction, density, etc., and obey simplified physics principles defined within the game's engine. Each block in the game can have multiple different shapes as well as being made from one of three materials (wood, ice or stone). Each bird is assigned one of five different types (red, blue, yellow, black or white). Each of these bird types are strong/weak against certain block materials, as well some types possessing secondary abilities which the player can activate during the bird's flight. The player can choose the angle and speed with which to fire a bird from the slingshot, as well as a tap time for when to activate the bird's special ability if it has one, but cannot alter the ordering of the birds or affect the level in any other way. Pigs are killed once they take enough damage from either the birds directly or by being hit with another object. The ground is usually flat but can vary in height for certain difficult levels. TNT can also be placed within a level and explodes when hit by another object. The difficulty of this game comes from predicting the physical consequences of actions taken, and accurately planning a sequence of shots that results in success. Points are awarded to the player once the level is solved based on the number of birds remaining and the total amount of damage caused.

\subsection{AIBIRDS level generation competition}

Whilst the AIBIRDS competition has been running annually for many years, a second track of the competition was started in 2016 known as the AIBIRDS level generation competition. This new competition revolves around developing procedural content generators (PCG) that can autonomously create Angry Birds levels. These generators must create levels that are both solvable and physically stable. Generated levels should also be fun and creatively designed, as well as providing the player with a suitable level of challenge. Currently the generators create levels for a clone version of Angry Birds, due to the fact that the real Angry Birds game is not open source, but recent engine improvements have allowed the generators to create levels very similar to those seen in the real Angry Birds game. In fact, several levels that were used in this year's AIBIRDS competition were converted from levels created by generators from the AIBIRDS level generation competition. Figure 2 provides an example generated level that was used in the AIBIRDS competition and was created using the algorithm described in \cite{togx5}. We also discuss later some ways in which both these competitions could be combined to increase the abilities and performance of agents, as well as helping to create better levels.

\begin{figure}
    \centering
  \includegraphics[width=1.0\linewidth]{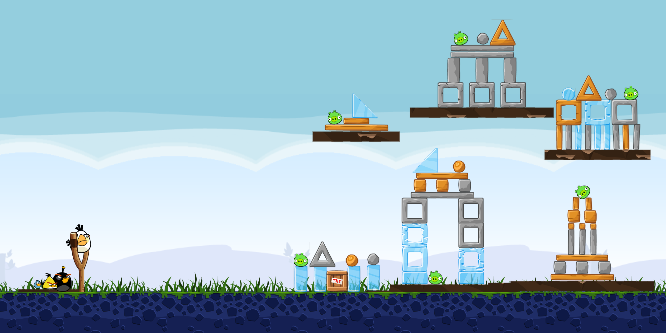}
  \caption{An example generated level that was used in this year's AIBIRDS competition.}
\end{figure}

\section{AIBIRDS Competition}

\subsection{Rules}

Developed agents can be written in any programming language, although we strongly recommend the use of Java to make integration with existing software easier. 
Each Angry Birds game instance is run on a game server while the agent itself is executed on a client computer, see Figure 3 for a server-client architecture diagram. 
Client computers have no access to the internet and can only communicate with the game server via the specified communication protocol. No communication with other agents is possible and each agent can only access files in its own directory. Each agent is able to obtain screenshots of the current Angry Birds game state from the server and can submit actions and other commands back to it. The game is played in SD mode and all screenshots have a resolution of 840x480 pixels. Agents that attempt to tamper with the competition settings or try to gain an unfair advantage will be disqualified.

\begin{figure}
    \centering
  \includegraphics[width=0.9\linewidth]{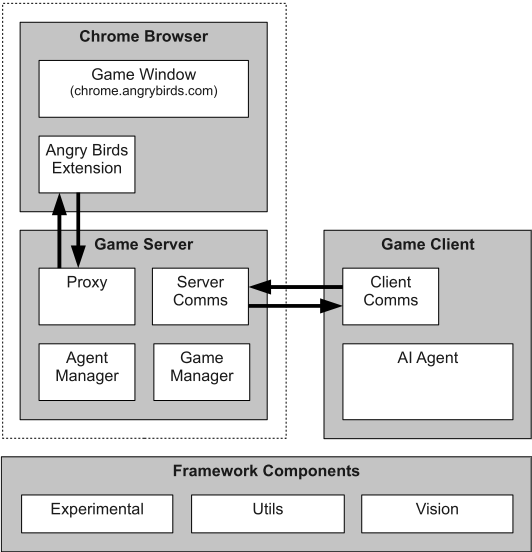}
  \caption{Server-client architecture.}
\end{figure}

The following objects are used for the competition levels: All objects, background, terrain, etc., that occur in the first 21 Poached Eggs levels of the Chrome version of Angry Birds. In addition, the competition levels may include the white bird, the black bird, TNT boxes, triangular blocks and hollow blocks. No other objects are used. The vision module of the provided game playing software recognises all relevant game objects and all birds and pigs, including the terrain but not the background. All competition levels use the same background that occurs in the first 21 Poached Eggs levels.





\subsection{Naive agent}

The source code for a naive agent is provided to all competition entrants as a useful starting point upon which to create their own agent. Objects within the level are first identified using a computer vision module, which converts the raw pixel image input into a easier to manage list of object types, sizes, materials and locations, see Figure 4. The naive agent also has an additional trajectory module, which calculates two possible release points, one firing horizontally (low trajectory) and the other vertically (high trajectory), that result in the current bird hitting a specified pig (assuming no objects are blocking the bird's trajectory). The naive agent always fires the currently selected bird at a randomly chosen pig using either a low or high trajectory (also chosen at random).  No other objects apart from the current bird and pigs are used when determining a suitable shot, and tap times are fixed for each bird based on the total length of its trajectory. This agent can therefore make shot calculations quickly and accurately but is unlikely to be skilled enough to solve more challenging Angry Birds levels. Participating teams are advised to use both the vision and trajectory modules provided with this agent, but will likely need to improve the techniques and strategies used to solve levels. Entrants are also provided with several other built-in functions. These include a function that identifies which blocks support a specific block, and a function which identifies whether particular trajectories to pigs are obstructed by other objects.


More detailed explanations about how the naive agent and server software works can be found on the AIBIRDS website \cite{agent3}. This website also provides instructions on how to begin coding your own intelligent agent for Angry Birds and has several available open source entries to try out.

\begin{figure}
    \centering
  \includegraphics[width=1.0\linewidth]{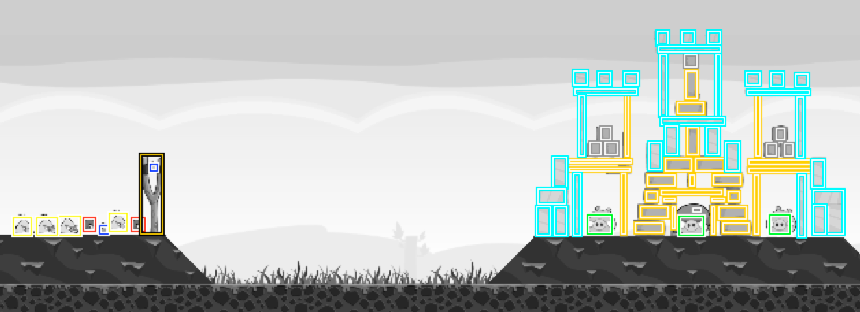}
  \caption{The example level from Figure 1, with blocks, pigs and birds identified using the computer vision module.}
\end{figure}

\subsection{Scoring and tournament procedure}

During the competition, there is a time limit to play a given set of Angry Birds levels automatically and without any human intervention. The competition is played over multiple knock-out rounds. In each round, the agents that achieve the highest combined game score over all solved levels proceed to the next round. The agent with the highest combined game score in the grand final is the winner of the competition. There is also an additional side competition in which the best agents from the main AI competition are pitted against human players attending the host conference.





\subsubsection{Main AI competition}

The main AI competition consists of three group rounds to determine the two best agents (qualification, quarter-finals and semi-final) as well as a final showdown to decide the overall winner (grand final). For each round we have a dynamically updated leaderboard where all agents are ranked according to their total score. All levels used in the competition are brand new and are not known in advance by any participating team. Agents have a total time of 30 minutes to solve eight levels in each round. Each of the eight game levels within a round can be accessed, played and replayed in any arbitrary order. After the 30 minute time limit for a round is reached, the connection between agents and the game server is terminated. Agents then have up to two minutes to store any information they wish to keep and to stop running. After two minutes the organisers terminate the agents if they are still running. Agents cannot be modified during the competition. 

The first round of the competition is a qualification round, where up to 16 teams are selected to proceed onto the next round. As there were only ten participating teams this year, this round was simply used to help divide up the teams into groups for the quarter-finals. For the quarter-finals we had one group of four teams and two groups of three teams, based on the scores achieved in the qualification round. Any team can query the current group high score for each level, but not the high scores of other groups. The top four teams across all three quarter-final groups move on to the semi-final. 
The final four teams that make it to the semi-final then all play in the same group. Any team can query the current high score for each level, and the two best teams qualify for the grand final. During the grand final both teams can query the current high score for each level, with the winner of this match being the 2017 AIBIRDS champion. The four semi-finalists also qualify for the man vs machine challenge.

\subsubsection{man vs. machine challenge}

During the man vs. machine challenge we test if the best agents can beat humans at playing Angry Birds. For this challenge we use four new Angry Birds levels not included in the main AI competition, that each player has 10 minutes to solve. Each game level can be accessed, played and replayed in any arbitrary order. Participating human players play the game levels first. Each player can participate only once. A leaderboard is kept which ranks the human players according to their overall score (sum of individual high scores per level). After the human players the four best agents (those which qualified for the semi-final in the main AI competition) are run in parallel on the same game levels with the same time limit. The player with the highest overall score, man or machine, wins this challenge.

\section{Competition Agents}

This year we had 10 agents submitted from teams across 12 countries. More information on the agents entered into both this year's and previous competitions are available on the AIBIRDS website \cite{agent2}.

\subsection{Datalab (Placed 7th; Czech Technical University in Prague; Czech Republic; First entered in 2014)}

The Datalab agent uses a combination of four different strategies when attempting to solve a level. These can be described as the destroy pigs, building, dynamite and round blocks strategies. The decision of which strategy to use is based on the environment, possible trajectories, currently selected bird and remaining birds. The destroy pigs strategy attempts to find a trajectory that intersects with as many pigs as possible. The building strategy identifies groups of connected blocks that either protect pigs or are near to them. The decision of which blocks within the building are suitable targets is based on its location, size, shape, material and relative placement within the structure, as well as the shape of the building itself. The shot that will cause the most damage to the building is then selected. The dynamite strategy ranks each TNT box within the level based on the number of pigs, stone blocks and other TNT boxes that are nearby. The round blocks strategy attempts to either hit round blocks directly or else destroy objects that are supporting round blocks. The tap time for each bird is fixed based on the location of the first obstacle in its trajectory, with the exception of the white bird.

\subsection{IHSEV (Placed 2nd; \'Ecole nationale d'ing\'enieurs de Brest; France; First entered in 2013)}

The IHSEV agent creates an internal Box2D simulation of the level, within which it tries out many shot angles and tap times. These mental simulations are carried out in parallel to identify the shot that destroys the most pigs. The simulation is not a perfect representation of the environment and great care is taken when perceiving and reconstructing each level. The vision module has also been slightly improved from the base code provided so that objects are more robustly identified. The agent does not use any information about the number or type of remaining birds when deciding which shot to take. A future plan to adapt the agent's environmental simulation based on the deviation between the actual and expected outcome of a shot was proposed but has not yet been implemented. This is currently the only agent that considers multiple different tap times for activating each bird's abilities successfully.

\subsection{Angry-HEX (Placed 3rd; Universit\`a della Calabria, Vienna University of Technology, Marmara University, Max Planck Institut fuer Informatik; Italy, Austria, Turkey, Germany; First entered in 2013)}

The Angry-HEX agent uses HEX programs to deal with decisions and reasoning, while the computations are performed by traditional programming. HEX programs are an extension of answer set programming (ASP) which use declarative knowledge bases for information representation and reasoning. The Reasoner module of this agent determines several possible shots based on different strategies. These shots are then simulated using an internal Box2D simulation, with the shot that kills the most pigs being selected as the ideal action. If the estimated number of killed pigs is the same for multiple possible shots, then the shot that also destroys the most objects is selected. 
The trajectory module of the base program was improved to take the thickness of the currently selected bird into account, as well as the ability to select several different points on a block as the target location. The tap time for each bird is fixed based on the location of the first obstacle in its trajectory, with the exception of the white bird. This agent can also remember the shots and strategies previously carried out, to aid them when re-attempting levels.

\subsection{Eagle's Wing (Placed 1st; University of Alberta and Zazzle Inc.; Canada; First entered in 2016)}

The Eagle's Wing agent chooses from five different strategies when deciding what shot to perform. These are defined as the pigshooter, TNT, most blocks, high round objects and bottom building blocks strategies. The decision of which strategy to use is based on the estimated utility of each approach with the currently selected bird. This utility is calculated based on the level's features and how these compare to a small collection of practice levels that are used to train the agent with the machine learning method xgboost. The pigshooter strategy attempts to find a trajectory that either targets an unprotected pig or includes multiple pigs within it. The TNT strategy aims for any TNT box that can cause significant damage to a large region. The many blocks strategy finds the trajectory that destroys the most blocks (highly dependent on the type of bird being used). The high round objects strategy attempts to destroy objects close to large round objects that are high above the ground, hopefully causing them to fall onto pigs. The bottom building block strategy targets blocks that are important to a structure's overall stability. The tap time for each bird is fixed based on the location of the first obstacle in its trajectory, with the exception of the white bird.


\subsection{s-birds (Placed 5th; Dhirubhai Ambani IICT; India; First entered in 2013)}

The s-birds agent has two different approaches for determining the most effective shot to perform. The first strategy is called the bottom-up approach and identifies a set of candidate target blocks for the level based on the potential number of affected pigs. The second strategy is called the top-down approach and utilizes the crushing/rolling effect of a bird or round block onto pigs, as well as the toppling effect of thinner blocks. Suitable target blocks are identified for each method and are then ranked based on the expected number of pigs killed and the likelihood of the shot's success. The penetration factor of specific bird types against certain materials is also considered when determining if a block can be hit. The tap time for each bird is fixed based on the total length of its trajectory, with the exception of the white bird.


\subsection{BamBirds (Placed 9th; Bamberg University; Germany; First entered in 2016)}

The Bambirds agent creates a qualitative representation of the level and then chooses one of nine different strategies based on its current state. This includes approaches such as utilizing blocks within the level to create a domino effect, targeting blocks that support heavy objects, maximum structure penetration and prioritizing protective blocks, as well as simpler options such as targeting pigs/TNT or utilizing certain bird's special abilities. These strategies are each given a score based on their estimated damage potential for the current bird type. A strategy is then chosen randomly, with this score being used to determine the likelihood of selection (i.e. shots that are believed to be the most effective are more likely to be chosen). The tap time for each bird is fixed based on the total length of its trajectory, with the exception of the white bird. This agent can also remember the shots and strategies previously carried out, to aid them when re-attempting levels.

\subsection{PlanA+ (Placed 4th; Sejong University; South Korea; First entered in 2014)}

The PlantA+ agent alternates between two different strategies each time it attempts a level. The first strategy involves identifying two possible trajectories to every pig and TNT within the level, and then counting the number of blocks (for each material) that are blocking each trajectory from being successful. The agent then compares the type of bird that is currently available against the number and material of blocks blocking each trajectory, to calculate a heuristic for each possible shot. This heuristic value defines the likelihood of the bird successfully making it to the specific target. The second strategy is similar to the first, except that the number of pixels crossing the trajectory is used rather than the number of blocks. Parameter values for each strategy are first identified by humans using trial and error, but are then optimised using a simple greedy-style algorithm. The tap time for each bird is fixed based on the location of the first obstacle in its trajectory, with the exception of the white bird.



\subsection{AngryBNU (Placed 10th; Beijing Normal University; China; First entered in 2017)}

The AngryBNU agent uses deep reinforcement learning, more specifically it uses deep deterministic policy gradients (DDPG), to build a model for predicting suitable shots in unknown levels based on generalised experience from previously played levels. DDPG is a combination of Deep Q-networks, Deterministic policy gradient algorithms and actor-critic methods, allowing for efficient deep learning in continuous action spaces, such as the environment present in Angry Birds. The model trained with DDPG can be used to predict optimal shot angles and tap times, based on the features within a level. The level features that are considered when training and utilising this model are the current bird type, the distance to the target points, and a 128x128 pixel matrix around each target (nearby objects). The level screenshot received by the agent is also transformed into an annotated image that retains relevant features while discarding those which are unnecessary. This pre-processing allows for more efficient generalisation when only a limited training set is available. Continuous Q-learning (SARSA) is used as the critic model and policy gradient is used as the actor model. By following this process, a deep learning model is trained on the original Angry Birds levels that are available, which allows the agent to predict the best target point for a shot based on the level's features. 

\subsection{Condor (Placed 8th; UTN Facultad Regional Santa Fe; Argentina; First entered in 2017)}

The Condor agent chooses from five different strategies when deciding what shot to perform. These are defined as the structure, boulder, TNT, bird and alone pig strategies. Each strategy has corresponding level requirements to decide whether it's considered or discarded for the current shot. Each strategy also has a numerical weighting based on human analysis of their potential impact for the current level. The structure strategy detects the shape of structures (two or more connected blocks) and classifies them as either a fortress or a lookout. Fortresses are targeted at the top left position, whilst lookouts are targeted at the mid-point. The boulder strategy targets round blocks next to pigs. The TNT strategy simply targets TNT boxes. The bird strategy identifies suitable blocks to hit based on their material and the type of bird that is currently available. The alone pig strategy targets pigs that are reachable and unprotected. The agent also uses an improved system when waiting for the resulting movement caused by a shot to finish, using a dynamic system rather than a static timer. The agent also re-adjusts the target point slightly if it believes this may give a better shot result (multiple target points for each object). The tap time for each bird is fixed based on the total length of its trajectory.



\subsection{Vale Fina 007 (Placed 6th; Technical University of Crete; Greece; First entered in 2017)}

The Vale Fina 007 agent uses reinforcement learning (specifically Q-learning) in an attempt to identify suitable shots for unknown levels based on past experience. In order to describe the current state of a level, a list of objects is used that contains information about every object within it. An object is described based on several features, including the object angle, object area, nearest pig distance, nearest round stone distance, the weight that the object supports, the impact that the current bird type has on the object, and several others. Q-learning is then used to associate the features of the objects within a level to certain actions (shots) that result in success. These features are weighted based on their perceived importance for a collection of sample testing levels. Unfortunately, no more information was provided about this agent, so a more in-depth comparison between this and the other reinforcement learning agent (AngryBNU) is not possible. 


\begin{table*}
\begin{center}
\begin{tabular}{|p{2.8cm}|p{2.0cm}|p{2.0cm}|p{2.0cm}|p{2.0cm}|}
    \hline
    \textbf{Agent} & \textbf{Qualification} & \textbf{Quarter-finals} & \textbf{Semi-final} & \textbf{Grand final}\\ \hline
        (1) \quad  Eagle's Wing & 416,650 & 175,510 & 350,900 & 355,700\\ \hline
        (2) \quad  IHSEV & 415,370 & 261,600 & 415,890 & 275,110\\ \hline
        (3) \quad  Angry-HEX & 405,340 & 242,980 & 238,040 & -\\ \hline
        (4) \quad  PlanA+ & 455,110 & 172,410 & 225,780 & -\\ \hline
        (5) \quad  s-birds & 155,980 & 147,120 & - & -\\ \hline
        (6) \quad  Vale Fina 007 & 332,630 & 106,930 & - & -\\ \hline
        (7) \quad  Datalab & 483,750 & 97,100 & - & -\\ \hline
        (8) \quad  Condor & 282,000 & 94,600 & - & -\\ \hline
        (9) \quad  BamBirds & 307,890 & 89,830 & - & - \\ \hline
        (10) \enspace  AngryBNU & 0 & 0 & - & -\\ \hline
  \end{tabular}
\caption{Round scores for each agent at the 2017 AIBIRDS competition (ordered based on final ranking)}
\end{center}
\end{table*}

\begin{table}
\begin{center}
\begin{tabular}{|p{2.0cm}|p{1.0cm}|p{1.0cm}|p{1.0cm}|p{1.0cm}|}
    \hline
    \textbf{Agent} & \textbf{2016} & \textbf{2015} & \textbf{2014} & \textbf{2013} \\ \hline
        Eagle's Wing & 5th & - & - & -\\ \hline
        IHSEV & 2nd & 4th & 4th & 8th\\ \hline
        Angry-HEX & 7th & 2nd & 7th & 4th \\ \hline
        PlanA+ & - & 5th & 3rd & -\\ \hline
        s-birds & 8th & 6th & 6th & 11th\\ \hline
        Vale Fina 007 & - & - & - & -\\ \hline
    	Datalab & 3rd & 1st & 1st & -\\ \hline
    	Condor & - & - & - & -\\ \hline
    BamBirds & 1st & - & - & -\\ \hline
    AngryBNU & - & - & - & -\\ \hline
  \end{tabular}
\caption{Agent rankings at previous AIBIRDS competitions}
\end{center}
\end{table}

\begin{table}
\begin{center}
\begin{tabular}{|p{1.7cm}|p{2.1cm}|p{2.1cm}|p{1.1cm}|}
    \hline
    \textbf{Agent} & \textbf{Benchmark set 1} & \textbf{Benchmark set 2} & \textbf{Total}\\ \hline
        Eagle's Wing & 941,840 & 896,630 & 1,838,470 \\ \hline
        IHSEV & 915,540 & 513,740 & 1,429,280 \\ \hline
        Angry-HEX & 865,470 & 668,690 & 1,534,160 \\ \hline
        PlanA+ & 922,480 & 653,720 & 1,576,200 \\ \hline
        s-birds & 732,080 & 223,710 & 955,790 \\ \hline
        Vale Fina 007 & 661,870 & 292,060 & 953,930 \\ \hline
    	Datalab & 947,240 & 1,060,610 & 2,007,850 \\ \hline
    	Condor & 765,870 & 190,860 & 956,730 \\ \hline
    BamBirds & 774,730 & 242,150 & 1,016,880 \\ \hline
    AngryBNU & 763,720 & 618,820 & 1,382,540 \\ \hline
    Naive & 855,370 & 584,290 & 1,439,660 \\ \hline
  \end{tabular}
\caption{Agent benchmark scores on original Angry Birds levels}
\end{center}
\end{table}

\section{Results}

During the competition agents played in repeated rounds of 8 levels which needed to be solved in 30 minutes, as per our already described tournament procedure. A total of 32 levels were created for the four rounds required. These levels were created using a variety of techniques, including both hand-designed levels written by the competition organisers, as well as generated levels from this year's companion AIBIRDS level generation competition. 
The additional man vs. machine challenge was held after the main AI competition once the final agent rankings were known. The final result for each agent in each round of the main AI competition is shown in Table I. A dash in this table indicates that the agent was eliminated due to a low score and did not proceed to this round of the competition.

Although the qualification round was only used to divide agents into smaller groups for the quarter-finals, it was still useful in identifying the agents that would likely perform best in the future. Previous two-time winner Datalab was the best scoring agent for this round and looked set to dominate the following rounds at well. However, disaster struck in the quarter-finals with Datalab being ranked 7th, well below the requisite 4th place ranking to make it into the semi-finals. Instead the quarter-final round, and the subsequent semi-final round, had their highest score achieved by IHSEV. Even though IHSEV performed best in both the quarter and semi-final rounds, and was a clear favourite going into the grand final, it ultimately lost to Eagle's Wing.

We can also compare the results from this year's competition against the agent rankings from past years, see Table II, as well as the benchmark scores for each agent, see Table III. These benchmark scores are the total scores for each agent, when given 120 minutes to solve each of the first two sets of ``poached eggs'' levels from the original Angry Birds game (21 levels in each benchmark set). These levels are available to participating teams before the competition, and allow us to compare each agent's performance when playing the levels it has been trained and fine-tuned on against the unknown levels of the competition.

For the man vs. machine challenge we had 45 human participants, facing the four best agents from the main AI competition (Eagle's Wing, IHSEV, Angry-HEX and PlanA+). The best performing agent was PlanA+ which solved three levels and got 73,540 points, second was IHSEV which also solved three levels for a total of 71,020 points, third was Eagle's Wing (the winner of the main AI competition) which solved two levels for 43,830 points, and last was Angry-HEX which solved one level for 22,920 points. Only one of the 45 human players was unable to beat all agents, proving once again that agents still have a long way to go to achieve a level of skill equivalent to that of a human. For the record, the best score achieved by a human player was 178,290 points by Sebastian Rudolph from the University of Dresden, Germany.

\section{Discussion}

\subsection{Competition issues}

Due to the fact that some of the agents decided to implement their own computer vision module, there were occasionally times when agents would fail to detect key objects such as birds or pigs and be unable to complete a level. As to not penalise agents with this problem too much, any agent that was stuck on a particular level for too long without making a shot had its level reset manually. This would often fix any vision issues, and if it didn't then the failure was on the agent. All levels were tested before the competition with our provided computer vision module, to ensure that there were no problems with agents that chose to use it. The deep learning techniques used by AngryBNU also required additional memory than was typically available in past competitions, so all client computers had their memory increased from 4GB to 8GB. Apart from these small complications, everything else in the competition's procedure went exactly to plan.



\subsection{Agent comparison}

While the overall performance of each agent during this year's competition should be clear from the results, there is much discussion to be had on why certain agents were more successful than others and what this means for future work around developing AI systems for physical reasoning in unknown environments.

\subsubsection{Agent techniques}

By looking at the techniques used by each agent, we can try and identify why this may lead to vastly different agent performances against certain types of levels. From our own observations, each of the 10 agents entered into this year's competition can be grouped into one of three main categories based on the AI approach they used. Heuristic-based agents (Datalab, Eagle's Wing, s-birds, BamBirds, PlanA+ and Condor), Simulation-based agents (IHSEV and Angry-HEX) and Reinforcement Learning agents (AngryBNU and ValeFina 007). Naturally there is some crossover between these groups, Angry-HEX for example uses heuristic calculations to identify shots that are worth simulating, but these categories allow us to discuss the different ways to approach this problem in broader and more general terms without having to refer to specific agents. 

Heuristic-based agents are by far the most varying in their performance, as they effectively choose from a fixed number of strategies based on their level observations. The skill of a heuristic-based agent is therefore entirely dependent on the skill of the human designer, and their ability to identify common methods for solving levels. These agents have traditionally performed very well in both this year's and past competitions, but struggle with levels that cannot be solved using one of their pre-defined strategies. 
Simulation-based agents do not suffer from this limitation as much, as they instead simulate a variety of different possible shots using an internal simulator and pick that which has the best outcome. This method typically takes longer than the simpler heuristic-based approaches, especially on big levels with lots of objects, but can often find solutions to more non-traditional level designs. The problem with this approach lies in the fact that the internal simulation used by these agents is not a perfect representation of the actual Angry Birds game engine and so its estimated shot results can sometimes be wildly inaccurate, leading to very strange and foolish shots. 
The last approach is that of reinforcement learning. This year was the first time that agents in the competition have used this technique, and unfortunately the results were far from groudbreaking. While advanced reinforcement learning techniques such as deep learning have proven successful in many other video games \cite{dl1}, they require a large number of varied training levels on which to practise (something which the version of Angry Birds we are using does not currently possess).


\subsubsection{Shot time}

As previously mentioned, not all agents take the same length of time to make a shot, with many taking much longer to consider different options before acting. This is an additional factor that must be considered in conjunction with the strategy being used. The timed nature of this competition means that simulation-based strategies, such as those used by IHSEV, may not be the best approach. A 2017 paper around the development of an Angry Birds hyper-agent based on the 2016 competition entries, found a moderate negative correlation between the average score and shot time
for each agent \cite{togx6}. This would suggest that having a faster shot time typically leads to a greater overall score, which is likely due to the increased number of level attempts this results in. Given more time to simulate a greater number of shot possibilities may allow IHSEV to perform better, although the discrepancy between its own internal simulation and that being used within the actual game would still hinder this approach. Allowing agents an unlimited or highly extended amount of time to make decisions would heavily restrict our real-world applicability however, as an AI system that can perfectly reason about its environment is pointless if the time required to make decisions is too great. This same paper also found that heuristic and simulation-based approaches performed better at certain levels, suggesting that there is no one-type-fits-all strategy.

\subsubsection{Meta strategies}

Apart from the techniques used by each agent to solve levels, another degree of complexity that this competition brings is that of meta-strategies for determining which specific levels to play. The time given to each agent to solve a round of the competition is typically high enough that each level can be attempted multiple times. Some agents choose to attempt all levels
once before replaying any unsolved levels (such as Datalab
and Eagle’s Wing) whilst others attempt a level multiple
times before moving on (such as s-birds).
Angry-HEX and Bambirds are also able to remember the
shots and strategies previously carried out, to aid them when
re-attempting levels later on. Whilst most agents try to solve
all levels before re-attempting those already solved, Bambirds
calculates a probability of attempting each level based
on an estimated number of points for solving it, the number
of times it has been played and the current score for that
level. Agents can also see the scores of other agents, to determine which levels its counterparts are struggling with. Whilst these strategies can be very influential on an agent's final score, they are not yet sufficiently complex to warrant a full game theory style investigation and currently have little bearing outside of the competition environment.

\subsubsection{Creative levels}

Several of the levels used in this year's competition were designed to require some form of creative reasoning to solve them. These levels typically require agents to make a non-obvious first shot in order to clear the level with the second shot, see Figure 5 for an example. Human players can easily tell that if they first destroy the wooden support blocks the stone blocks will fall and leave the pig exposed, yet not one agent was able to solve this seemingly simple level. 
Most agents always targeted seemingly important objects such as Pigs or TNT, as well as making greedy shots with the current bird without considering those still to come.
Planning more than one shot ahead may seem like an incredibly difficult task without knowing the outcome of the initial shot, but humans who play Angry Birds can still come up with intelligent shot sequences without this information. It seems as though human intuition about how physical environments will react to certain actions can extend multiple steps into the future. Levels that are designed to deliberately exploit agent limitations and biases, demonstrate the need to develop and combine different AI techniques across multiple fields in order to achieve success. 

\begin{figure}
    \centering
  \includegraphics[width=1.0\linewidth]{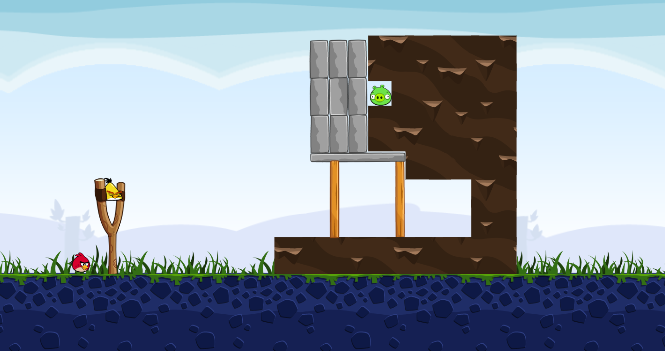}
  \caption{A level used in the quarter-finals round that requires agents to plan multiple shots forward to achieve success.}
\end{figure}

\subsubsection{Previous competitions}

Comparing the rankings of this year's agents against those same agent's rankings in previous competitions, see Table II, we can see that there is a large amount of variation in each agent's final ranking from year to year. Previous two-time winner Datalab, and last year's winner BamBirds did surprisingly poorly in this year's competition, whilst this year's winner Eagle's Wing rose from 5th place last year. This is likely caused by the nature of our competition's ``multiple round" design, which has been structured in such a way to give weaker agents a greater chance of winning than if the sum of each agent's scores across all 32 competition levels was used. This format means that factors such as the decision of which levels to use for each round is very important. Swapping the levels used in the semi-final and grand final rounds may have resulted in a completely different competition winner. Nevertheless, the fact that certain agents perform better in different rounds with different levels indicates that the variety of Angry Birds level designs used in the competition is sufficiently diverse such that no agent is currently skilled enough to successfully solve them all, and definitely not to a human's level of performance.

\subsubsection{Agent benchmarks}

Whilst the competition's levels are unknown, the benchmark levels are provided to participating teams beforehand as a way to test and evaluate their agent(s). By comparing each agent's scores for these levels against those in the competition, we can see that some agents relied much more heavily on the design of the known levels than others. The clearest example would be AngryBNU which did reasonably well on the benchmark levels, although not as good as some other agents, but failed to score any points at all during the competition. This would suggest that the deep reinforcement learning techniques it uses have been trained too heavily on the benchmark levels (i.e. overfitting), and thus cannot successfully adapt to the previously unseen levels used in the competition. As will be discussed later, recent improvements in the development of Angry Birds level generators could help alleviate this problem by providing a much larger selection of training levels.


\subsubsection{Man vs. Machine}

In order to estimate how close we are to achieving our goal of agents with human level performance, we compared the skill of the best four agents against human participants in the man vs. machine challenge. In previous iterations, humans have always won with a wide,
but shrinking margin. In 2013, half of human participants
were better than the best AI, while in 2014 it was a third, and in 2015 / 2016 the winning agent ended up being among the best eighth of
all human players who participated. To up the stakes this year, we significantly increased the complexity and difficulty of the levels used in this challenge. 
As a result, the overall performance of our agents against human players dropped dramatically compared to previous years. This suggests that while agents have been improving in their ability to play Angry Birds levels with a more traditional design, those that require creative reasoning to solve them still pose a significant challenge.

\subsection{Combining AIBIRDS competitions}

There are several ways in which the two current AIBIRDS competitions (agent and level generation) could be combined. As previously mentioned and utilised in this year's competition, level generators can be used to create additional levels for testing and evaluating the performance of an agent beyond the original hand-designed levels that the game currently provides. This ability to rapidly create new and unknown levels means that it is now possible to construct a large database of training levels for agents focussed around using reinforcement learning techniques, such as AngryBNU or Vale Fina 007. This addition could dramatically improve the performance of agents that use these techniques, particularly as they performed so poorly this year compared to other more traditional AI approaches.
Agents are also extremely useful in the AIBIRDS level generation competition, as they can be used to evaluate and test levels created by the generators. Different agents could also be used to test a generated level against different playstyles, or to determine whether a level is too hard or too easy based on the number of agents that can solve it and how long it takes them. 
We also hope to be able to link both the AIBIRDS agent and level generation competitions in the future, perhaps with agents trying to beat generated levels and generators trying to create levels that are difficult for agents.




\subsection{Research, education and teaching}

The AIBIRDS website \cite{new88} holds an extensive repository of resources for anyone wishing to enter the competition or conduct research around it. This includes open source code and papers on prior agents to assist newcomers, benchmarking software for comparing different techniques, and extensive details on past competitions. There are a wide range of techniques that are yet to be successfully implemented which could dramatically increase agent performance, ranging from incredibly complex machine learning algorithms to simply better heuristics for evaluating shots. We hope that this paper will inspire others to take up the challenge, perhaps you could be the one who finally cracks this problem and develops a skilful agent that can outperform humans.  







An additional version of the basic game-playing software has also been developed using the simple visual programming language Snap!. This version of the framework allows anyone to develop their own Angry Birds agent that can utilise multiple different strategies with little to no prior programming experience. We hope that this software will be used to promote computer science and AI to school children, whilst also learning basic programming skills. Separate AIBIRDS competitions focused on comparing agents developed by students from different countries could even be held, inspiring future generations of computer science researchers to take on the AI challenges of tomorrow.

\subsection{Future ideas}

We believe that there are several key areas where agents could be improved to help achieve better performance in the future. An overreliance on traditional Angry Birds levels seems to be a key weakness for a lot of agents, not just those that use reinforcement learning. While an increased number of available generated levels would help address this issue, it fails to tackle the underlying problem itself. Real-time machine learning techniques will need to be employed in order for agents to experiment and develop solutions to creatively designed levels. These agents can use human knowledge and insight as a useful starting point, but should also construct their own strategies independent of designer bias. It is clear from the results of this competition that this work still has a long way to go, and that any successful agent will need to combine multiple AI approaches to even stand a chance of rivalling human players.

The main improvements that could be made to future competitions would be to provide a greater variety of available levels, allowing competition entrants to better evaluate their agents before the actual competition, and to deliberately design levels that are meant to be difficult for agents but easy for human players, which would greatly help in identifying where certain AI approaches are lacking.
It would also be very helpful if more competition participants made their agents open source or provided more detailed information about their inner workings. While a competitive mentality and desire for teams with successful agents to keep their secrets to themselves is understandable, the main goal of this competition is to further the development and research around of agents that can interact within a physical environment. The more transparent participating teams are with their breakthroughs, the more successful future agents will be. Several previous competition entrants have already published their research and agent designs in academic conference or journal papers \cite{cite3,cite5,cite4,cite10,cite11,cite7,cite6}, and we hope that future participants will continue doing this.

Beyond the competition itself, the results and scores for each agent can help identify where the current techniques for physical reasoning in unknown environments are lacking. It would seem that even advanced AI techniques such as deep learning are not enough on their own to solve this problem. 
Humans can accomplish these tasks with little cognitive effort, but it would be incredibly difficult for an AI system to accomplish. Developing Angry Birds agents is only the start of this research but represents a clear and well-formed step in the right direction.
The AIBIRDS competition was recently investigated in a 2016 expert survey on progress in AI, which predicted that Angry Birds agents should be able to outperform humans in the next three years (median of all expert's predicted times) \cite{aifuture}. While we would be thrilled if such an agent was created in the next three years, we feel that this is a severe underestimation of the challenges and complexities involved in such an accomplishment.




\section{Conclusion}

In this paper we have presented an overview of the sixth AIBIRDS competition. The task of solving unknown Angry Birds levels posed by this competition is hugely relevant to many real-world problems that require physical reasoning. 
Even though many different AI approaches have been implemented to tackle this challenge, it appears that the problem is too difficult to be solved by any single technique alone. This year's competition even featured agents that attempted to use modern machine learning techniques, but unfortunately with little success.  
It seems evident from this and previous year's results that for future agents to succeed, they must draw from multiple areas of AI.
We hope that in the future many of the competition entrants will mutually share information about their techniques, further pushing forward towards our goal of developing an AI system that can reason and act within unknown physical environments based solely on visual inputs or other forms of perception. 
We would also like to thank the members of the IJCAI committee, competition entrants and all man vs. machine participants for their contribution to making this event possible. We intend to run this competition again in 2018 and encourage all interested teams to participate in this exciting challenge. 

\ifCLASSOPTIONcaptionsoff
  \newpage
\fi



\bibliographystyle{IEEEtran}
\bibliography{refs}
\end{document}